\documentclass[letterpaper, 10 pt, conference]{ieeeconf}  

\usepackage{fix-cm}
\usepackage{etex}

\usepackage{dblfloatfix}

\usepackage{nag}

\makeatletter
\@ifpackageloaded{xcolor}{}{%
\usepackage[table,x11names,dvipsnames,svgnames]{xcolor}%
}
\makeatother

\usepackage{colortbl}

\usepackage{graphicx}
\graphicspath{{./figures/}}

\usepackage{wrapfig}

\definecolorset{RGB}{lyft}{}{Red,194,39,36;Sunset,202,53,33;Orange,205,68,20;Amber,200,117,42;Yellow,242,169,52;Citron,186,188,44;Lime,112,159,33;Green,56,139,31;Mint,45,118,56;Teal,52,133,135;Cyan,60,132,202;Blue,55,94,248;Indigo,64,13,247;Purple,115,42,248;Pink,176,25,145;Rose,176,32,75}

\definecolorset{HTML}{h}{}{grapefruit,ED5565;bittersweet,FC6E51;sunflower,FFCE54;grass,A0D468;mint,48CFAD;aqua,4FC1E9;bluejeans,5D9CEC;lavender,AC92EC;pinkrose,EC87C0;lightgray,F5F7FA;mediumgray,CCD1D9;darkgray,656D78}

\definecolorset{RGB}{p}{}{ivory,221,221,221;navy,046,037,133;pine,051,117,056;aqua,093,168,153;azure,148,203,236;sand,220,205,125;peach,194,106,119;carnation,159,074,150;mulberry,126,041,084}

\definecolorset{RGB}{o}{}{black,000,000,000;mint,000,158,115;jeans,000,114,178;azure,086,180,233;banana,240,228,066;orange,230,159,000;ember,213,094,000;amaranth,204,121,167}

\usepackage{cite}

\usepackage{microtype}

\usepackage[american]{babel}
\usepackage{array}
\usepackage{multirow}
\usepackage{booktabs}
\usepackage{makecell} 

\ifcsname labelindent\endcsname
\let\labelindent\relax
\fi
\usepackage[inline]{enumitem}

\usepackage{subfig}

\newenvironment{lenumerate}[2][]
{\begin{enumerate}[label=(#2\arabic*),leftmargin=0.2in,itemindent=0.15in,#1]}
{\end{enumerate}}

\setlist*[enumerate,1]{label={\itshape\arabic*)}}

\makeatletter
\newcommand{\paragraphswithstop}{%
\let\copyparagraph\paragraph%
\renewcommand\paragraph[1]{\copyparagraph{##1.}}%
}
\makeatother

\usepackage[framemethod=tikz]{mdframed}

\makeatletter
\def\namedlabel#1#2{\begingroup
  #2%
  \def\@currentlabel{#2}%
  \phantomsection\label{#1}\endgroup
}
\makeatother

\makeatletter
\def\namedlabelphantom#1#2{\begingroup
  \def\@currentlabel{#2}%
  \phantomsection\label{#1}\endgroup
}
\makeatother

\newcommand{\parunskip}{\bgroup\unskip\parfillskip=0pt \par\egroup}

\makeatletter
\@ifpackageloaded{amsmath}{}{%
  \usepackage{amsmath}%
}
\makeatother
\usepackage{mathtools}
\usepackage{amssymb,amsfonts}
\makeatletter
\@ifundefined{proof}{%
  \usepackage{amsthm}%
}{}
\makeatother
\usepackage{gensymb}
\usepackage{mathtools}
\usepackage{bm}

\usepackage{refcount}
\newcommand{\newtheoremcontinued}[2]{%
  \newtheorem{#1}{#2}
  \newcounter{#1previous}
  \NewDocumentEnvironment{#1continued}{O{Continued} m}
  {
    \setcounter{#1previous}{\value{#1}}%
    \edef\tempcount#1{\getrefnumber{##2}}%
    \setcounter{#1}{\tempcount#1}%
    \addtocounter{#1}{-1}
    \begin{#1}[##1]
    }
    {
    \end{#1}
    \setcounter{#1}{\value{#1previous}}%
  }
}

\makeatletter
\@ifundefined{theorem}{
  \newtheoremcontinued{theorem}{Theorem}
  \newtheoremcontinued{corollary}{Corollary}
  \newtheoremcontinued{proposition}{Proposition}
  \newtheoremcontinued{lemma}{Lemma}
  \newtheoremcontinued{fact}{Fact}
  \newtheoremcontinued{problem}{Problem}
  \newtheoremcontinued{assumption}{Assumption}
  \newtheoremcontinued{property}{Property}
  \newtheoremcontinued{remark}{Remark}
  \newtheoremcontinued{example}{Example}
}{}
\@ifundefined{definition}{%
  \newtheoremcontinued{definition}{Definition}%
}
\makeatother

\usepackage{units}

  \newcommand{\newcolorlabel}[2]{%
  \expandafter\newcommand\csname #1\endcsname[1]{%
    \tikz[baseline]{\node[text=white,fill=#2,anchor=base,text height=1.3ex,text depth=0.1ex,font=\sffamily\bfseries]{##1}}}%
}

\newcommand{\newcommenter}[2]{%
  \expandafter\newcommand\csname #1\endcsname[1]{%
    \fcolorbox{#2}{#2}{\color{white}\textsf{\textbf{#1}}}
    {\color{#2}##1}}%
  \expandafter\newcommand\csname #1p\endcsname[1]{%
    \pdfcomment[color=#2,voffset=1Em]{#1:##1}}%
  \expandafter\newcommand\csname at#1\endcsname{%
    \fcolorbox{#2}{#2}{\color{white}\textsf{\textbf{@#1}}}
    {\color{#2}}}%
  \expandafter\newcommand\csname #1cite\endcsname[1]{%
    \csname #1\endcsname {[##1]}
  }%
  \expandafter\newcommand\csname #1ref\endcsname[1]{%
    \csname #1\endcsname {$\blacktriangleright$##1}
  }%
  \expandafter\newcommand\csname #1hl\endcsname[2]{%
    \colorbox{#2}{\color{white}\textsf{\textbf{#1}}}\sethlcolor{Azure2}\hl{##2}~%
    \expandafter\ifx\csname commentarrow\endcsname\relax$\leftarrow$\else \commentarrow[#2]\fi~%
    {\color{#2}##1}}%
  \expandafter\newcommand\csname #1hlp\endcsname[2]{%
    \sethlcolor{#2!50!white}\hl{##2}%
    \pdfmargincomment[color=#2,voffset=1Em,icon=Note,open=true,author=#1]{#1:##1}}%
  \expandafter\newcommand\csname #1st\endcsname[2]{%
    \colorbox{#2}{\color{white}\textsf{\textbf{#1}}}\sout{##2}~%
    \expandafter\ifx\csname commentarrow\endcsname\relax$\leftarrow$\else \commentarrow[#2]\fi~%
    {\color{#2}##1}}%
}
\newcommenter{TODO}{DodgerBlue1}
\newcommenter{rtron}{OliveDrab2}

\usepackage{comment}

\usepackage{pdfcomment}

\usepackage{soul}

\usepackage[normalem]{ulem}

\usepackage{csquotes}

\usepackage{listings}
\usepackage{suffix}

\usepackage{environ}

\makeatletter
\newsavebox{\boxifnotempty}
\newcommand{\displayifnotempty}[3]{\sbox\boxifnotempty{#2}\setbox0=\hbox{\usebox{\boxifnotempty}\unskip}%
  \ifdim\wd0=0pt
  \else
  #1\usebox{\boxifnotempty}#3%
  \fi%
}

\newcommand{\ifempty}[2]{\setbox0=\hbox{#1\unskip}%
  \ifdim\wd0=0pt%
  #2%
  \fi%
}

\newcommand{\ifnotempty}[2]{\setbox0=\hbox{#1\unskip}%
  \ifdim\wd0>0pt%
  #2%
  \fi%
}
\makeatother

\newcommand{\switchifempty}[3]{\sbox\boxifnotempty{#1}\setbox0=\hbox{\usebox{\boxifnotempty}\unskip}%
  \ifdim\wd0=0pt{}%
  #2%
  \else{}%
  #3%
  \usebox{\boxifnotempty}%
  \fi%
}

\makeatletter
\@ifundefined{chapter}{\usepackage{algorithm}}{\usepackage[chapter]{algorithm}}
\makeatother
\usepackage{algorithmicx}
\usepackage{algpseudocode}
\makeatother%

\usepackage{scrlfile}

\makeatletter
\newcommand*\newstoreddef[1]{
  \BeforeClosingMainAux{%
    \immediate\write\@auxout{%
      \string\restoredef{#1}{\csname #1\endcsname}%
    }%
  }%
}
\newcommand*{\restoredef}[2]{
  \expandafter\gdef\csname stored@#1\endcsname{#2}%
}
\newcommand*{\storeddef}[1]{
  \@ifundefined{stored@#1}{0}{\csname stored@#1\endcsname}%
}
\makeatother

\usepackage{pageslts}
\NewEnviron{tee}{\BODY\typeout{Marker Tee [start] ^^J \BODY ^^JMaker Tee [end]}}

\usepackage{cleveref}
\crefname{property}{property}{properties}
\Crefname{property}{Property}{Properties}
\crefname{assumption}{assumption}{assumptions}
\Crefname{assumption}{Assumption}{Assumptions}
\crefname{problem}{problem}{problems}
\Crefname{problem}{Problem}{Problems}
\crefname{fact}{fact}{facts}
\Crefname{fact}{Fact}{Facts}
\crefname{remark}{remark}{remarks}
\Crefname{remark}{Remark}{Remarks}
\usepackage{nameref}
\makeatletter
\@ifundefined{corollary}{}{
  \crefname{corollary}{corollary}{corollary}
  \Crefname{corollary}{Corollary}{Corollaries}
  \crefname{example}{example}{examples}
  \Crefname{example}{Example}{Examples}
  \crefname{remark}{remark}{remarks}
  \Crefname{remark}{Remark}{Remarks}
  \crefname{property}{property}{properties}
  \Crefname{property}{Property}{Properties}
  \crefname{assumption}{assumption}{assumptions}
  \Crefname{assumption}{Assumption}{Assumptions}
  \crefname{problem}{problem}{problems}
  \Crefname{problem}{Problem}{Problems}
  \crefname{fact}{fact}{facts}
  \Crefname{fact}{Fact}{Facts}
  \crefname{lemma}{lemma}{lemmas}
  \Crefname{lemma}{Lemma}{Lemmas}
  \crefname{proposition}{proposition}{propositions}
  \Crefname{proposition}{Proposition}{Propositions}
}

\usepackage{tikz}
\usetikzlibrary{calc}
\usetikzlibrary{matrix}
\usetikzlibrary{chains,scopes}
\usetikzlibrary{shapes.geometric}
\usetikzlibrary{arrows.meta}
\usetikzlibrary{decorations.markings}
\usetikzlibrary{decorations.pathreplacing}
\usetikzlibrary{backgrounds}

\tikzset{
  dim above/.style={to path={\pgfextra{
        \pgfinterruptpath
        \draw[>=latex,|->|] let
        \p1=($(\tikztostart)!1.5em!90:(\tikztotarget)$),
        \p2=($(\tikztotarget)!1.5em!-90:(\tikztostart)$)
        in(\p1) -- (\p2) node[pos=.5,sloped,above]{#1};
        \endpgfinterruptpath
      }
    }
  },
  dim double above/.style={to path={\pgfextra{
        \pgfinterruptpath
        \draw[>=latex,|->|] let
        \p1=($(\tikztostart)!3em!90:(\tikztotarget)$),
        \p2=($(\tikztotarget)!3em!-90:(\tikztostart)$)
        in(\p1) -- (\p2) node[pos=.5,sloped,above]{#1};
        \endpgfinterruptpath
      }
    }
  },
  dim below/.style={to path={\pgfextra{
        \pgfinterruptpath
        \draw[>=latex,|->|] let
        \p1=($(\tikztostart)!-1em!-90:(\tikztotarget)$),
        \p2=($(\tikztotarget)!-1em!90:(\tikztostart)$)
        in (\p1) -- (\p2) node[pos=.5,sloped,below]{#1};
        \endpgfinterruptpath
      }
    }
  },
}

\tikzset{
    right angle quadrant/.code={
        \pgfmathsetmacro\quadranta{{1,1,-1,-1}[#1-1]}     
        \pgfmathsetmacro\quadrantb{{1,-1,-1,1}[#1-1]}},
    right angle quadrant=1, 
    right angle length/.code={\def\rightanglelength{#1}},   
    right angle length=2ex, 
    right angle symbol/.style n args={3}{
        insert path={
            let \p0 = ($(#1)!(#3)!(#2)$) in     
                let \p1 = ($(\p0)!\quadranta*\rightanglelength!(#3)$), 
                \p2 = ($(\p0)!\quadrantb*\rightanglelength!(#2)$) in 
                let \p3 = ($(\p1)+(\p2)-(\p0)$) in  
            (\p1) -- (\p3) -- (\p2)
        }
    }
}

\newcommand{\pgfextractangle}[3]{%
    \pgfmathanglebetweenpoints{\pgfpointanchor{#2}{center}}
                              {\pgfpointanchor{#3}{center}}
    \global\let#1\pgfmathresult
}

\usetikzlibrary{shapes.arrows}
\newcommand{\commentarrow}[1][Azure4]{\tikz[baseline=-3pt]{\node[shape border uses incircle, fill=#1,rotate=180,single arrow, inner sep=1pt, minimum size=6pt, single arrow head extend=2pt]{};}}

\tikzset{ax/.style={-latex,line width=2pt}}

\tikzset{camera/.style={fill=Sienna1,fill opacity=0.5},%
image plane/.style={draw=RoyalBlue3,line width=2pt}}

\IEEEoverridecommandlockouts                              
\title{\LARGE \bf
Data-Driven Trajectory Prediction via Probabilistic Modes and Constrained Optimization
}

\author{Danyang Li$^{1}$, Ana Theodora Balaci$^{1}$, Matthew Cleaveland$^{2}$, and Roberto Tron$^{1}$
\thanks{$^{1}$Danyang Li, Ana Theodora Balaci, and Roberto Tron are with Boston University.
        {\tt\small \{danyangl, anab, tron\}@bu.edu},    
        $^{2}$Matthew Cleaveland is with MIT Lincoln Laboratory, Lexington, MA, 02421, USA.
        {\tt\small \{matthew.cleaveland\}@ll.mit.edu}.     
        }
\thanks{DISTRIBUTION STATEMENT A. Approved for public release. Distribution is unlimited.
This material is based upon work supported by the Department of the Army under Air Force Contract No. FA8702-15-D-0001 or FA8702-25-D-B002. Any opinions, findings, conclusions or recommendations expressed in this material are those of the author(s) and do not necessarily reflect the views of the Department of the Army.
© 2026 Massachusetts Institute of Technology.
Delivered to the U.S. Government with Unlimited Rights, as defined in DFARS Part 252.227-7013 or 7014 (Feb 2014). Notwithstanding any copyright notice, U.S. Government rights in this work are defined by DFARS 252.227-7013 or DFARS 252.227-7014 as detailed above. Use of this work other than as specifically authorized by the U.S. Government may violate any copyrights that exist in this work.
}%
}

\usepackage{tikz}
\usetikzlibrary{shapes.geometric, arrows.meta, fit, positioning, backgrounds, shapes, positioning}

\newcommenter{danyang}{red}
\newcommenter{mingyu}{DodgerBlue2}
\newcommenter{mcleav}{green}
\newcommenter{Ana}{orange}

\begin{document}

\maketitle
\thispagestyle{empty}
\pagestyle{empty}

\NewDocumentCommand\method{}{ProMo-CoOp}
\WithSuffix\NewDocumentCommand\method*{}{Probabilistic Modes and Constrained Optimization trajectory prediction}


\begin{abstract}
Predicting the future trajectories of surrounding agents is a fundamental requirement for safe autonomous driving. Since agent intent is inherently uncertain, multimodal prediction is essential to account for the full range of plausible future behaviors. We introduce \method{} (\method*), a two-stage data-driven framework. At design time, trajectories are clustered into behavioral modes and a Gaussian Mixture Model (GMM) provides probabilistic mode assignments. At prediction time, a candidate trajectory is generated for each mode by solving a convex optimization problem that enforces consistency with the observed trajectory and uses the precomputed mode assignments to guide prediction toward the selected behavioral mode. The framework also supports Signal Temporal Logic (STL)-derived behavioral constraints, with predicate satisfaction probabilities estimated using Conditional Flow Matching (CFM). Experiments show that \method{} outperforms state-of-the-art baselines, and that STL-based constraints can further improve performance when the predicted predicate satisfactions are reliable.


\end{abstract}

\section{Introduction}
Trajectory prediction is critical for autonomous driving, where surrounding agents may exhibit multiple plausible future behaviors~\cite{Alahi_2016_CVPR,rudenko2020human}. Since their underlying intent is unknown and may not be determined from the observed motion history alone, Multimodal Trajectory Prediction (MTP) aims to generate a set of plausible future trajectories rather than committing to a single prediction.



Existing MTP methods often fall short in at least one of three critical areas:
\begin{enumerate*}[label=\it{(\roman*)}]
\item \emph{coverage} (predictions span the entire space of likely outcomes);
\item \emph{feasibility} (predictions adhere to physical rules, kinematic limits, and social norms); or
\item \emph{explainability} (predictions correspond to a discernible intent or environment condition).
\end{enumerate*}
For autonomous driving, these three requirements are all consequential: insufficient coverage can miss the true future behavior, lack of feasibility can produce physically implausible or unsafe predictions, and lack of explainability makes it difficult to interpret or trust the predicted modes in downstream decision-making. Purely generative methods tend to emphasize coverage, but it is often difficult to ensure that their predictions remain feasible or correspond to an interpretable behavioral mode. By contrast, rule- or planning-based methods typically enforce feasibility, but often rely on hand-crafted assumptions that do not capture the full diversity of behaviors observed in data. Designing methods that simultaneously address all three requirements therefore remains challenging.

To bridge this gap, we propose \method{} (\method*). We identify a set of motion patterns directly from data through clustering and fit a Gaussian Mixture Model (GMM) to compute probabilistic mode assignments for each training sample; at prediction time, future trajectories are generated by an optimization problem conditioned on the selected pattern and guided by the precomputed mode assignments, thereby providing \emph{coverage} over the space of likely agent behaviors. The optimization is formulated as a convex program that guarantees each predicted trajectory lies within the convex hull of the training trajectories, which improves \emph{feasibility} by constraining predictions toward demonstrated behaviors. We further propose an extension of our method that incorporates Signal Temporal Logic (STL)-derived behavioral side information ~\cite{maler2004monitoring, fainekos2009robustness} directly into the optimization, through soft predicate-satisfaction constraints. STL is a powerful formal language designed to specify properties over real-valued signals, making it well suited to describing vehicle behaviors over time. This extension preserves the modularity of the base method while allowing \emph{explainability} of agent behavioral intent representations that go beyond purely statistical mode assignment.



Our \textbf{Contributions} are as follows:
\begin{enumerate}[label=\it{(\alph*)}, leftmargin=*, itemsep=0.15em, topsep=0.2em]
    \item We propose a data-driven framework for multimodal trajectory prediction that separates behavioral mode identification from trajectory generation. The probabilistic mode decomposition improves \emph{coverage} by representing multiple plausible futures and improves \emph{explainability} by making the predicted behavioral modes explicit, while the convex optimization step improves \emph{feasibility} by generating trajectories consistent with both the observed history and demonstrated trajectories;

    \item We extend our optimization framework to incorporate additional behavioral side information, including STL-based constraints, without modifying the underlying \method{} method;

    \item We show experimentally that the resulting base method is competitive with strong multimodal baselines, and that additional behavioral constraints can further improve performance when such side information is reliable.
\end{enumerate}

\section {Related Work}
Prior work in trajectory prediction can be viewed through the lens of the three challenges highlighted in the introduction: \emph{coverage}, \emph{feasibility}, and \emph{explainability}. A basic distinction is between unimodal and multimodal predictors~\cite{madjid2025trajectory}. Unimodal models generate a single output, typically a deterministic trajectory sequence or the parameters of a unimodal probability distribution~\cite{8317913,8672889,9157331}, and therefore do not explicitly address multimodal uncertainty. To better capture this uncertainty, multimodal approaches (such as \method) aim to produce diverse future behaviors.

A wide range of approaches has been proposed for multimodal trajectory prediction~\cite{huang2023multimodal}. Early work employed Gaussian Mixture Models (GMMs) for distribution modeling~\cite{wiest2012probabilistic}, but these models often produced unrealistic predictions. More recent methods based on Variational Autoencoders (VAEs)~\cite{lee2017desire}, Generative Adversarial Networks (GANs)~\cite{gupta2018social}, and Transformers~\cite{liu2021multimodal,9878832} have substantially improved expressive power and multimodal coverage. However, these models often provide limited control over feasibility, and their latent representations are not always easily connected to an explainable behavioral mode. Moreover, they can be sensitive to training instability, in which the optimization process tends to collapse toward a single dominant mode, producing predictions that lack diversity.

A second line of work seeks to improve feasibility, and in some cases explainability, by introducing planning structure into trajectory prediction (i.e., incorporating explicit structural knowledge beyond the observed motion history)~\cite{6696982,8793568,pmlr-v100-chai20a,9157523,9812107,song2022learning}. These approaches extend beyond purely history-based prediction by explicitly modeling the assumption that agents follow latent plans or goals. They are generally divided into two subcategories: model-based planning and learning-based planning. Model-based methods~\cite{6696982,8793568} improve feasibility and interpretability by integrating explicit physical principles and behavioral rules, but they typically lack adaptability to complex and diverse behaviors that are not easily encoded in hand-crafted rules. Learning-based methods~\cite{pmlr-v100-chai20a,9157523,zhao2021tnt,9812107} excel at inferring plans or intentions directly from observed data, but often sacrifice explainability and do not naturally enforce physical limits, safety constraints, or common-sense behavioral rules.

Recent hybrid methods attempt to combine the strengths of learning-based and model-based prediction. PRIME~\cite{song2022learning}, for example, generates representative candidate futures, also referred to as trajectory anchors, that satisfy environmental and kinematic constraints. However, PRIME relies on depth-first search to enumerate candidate behaviors, which may limit coverage due to its dependence on cutoff depth. Furthermore, it selects the final trajectory using learned scores from a neural evaluator, thereby lacking an explicit and interpretable criterion for intent selection. Like these hybrid approaches, our method combines data-driven behavior modeling with constraint-based trajectory generation. It differs, however, in that mode discovery is performed through a probabilistic decomposition of trajectory data, yielding an explicit representation of plausible behavioral modes that improves both coverage and explainability, while trajectory generation is carried out through a constrained optimization step that enforces consistency with both the observed history and the selected mode, thereby improving feasibility.

STL has also been used in related autonomous-driving settings, typically as an externally specified rule layer rather than a data-driven representation of behavior. For example, prior work has used STL robustness to score safety and right-of-way priorities in trajectory prediction~\cite{shi2025rulenet}, and has encoded traffic regulations as temporal-logic-style specifications for vehicle planning and control~\cite{shi2025digitizing,yu2026stlcontrol}. In contrast, our extension uses STL to represent behavioral structure associated with the discovered motion patterns and incorporates it directly into trajectory generation.

\section{Preliminaries}\label{sec:prelim}

\subsection{Notation}
We denote the state of an agent at time $t$ as $s_t = [x_t, y_t]^\top \in \mathbb{R}^2$, representing its two-dimensional position. A trajectory over the interval $[t_0, t_1]$ is written as $s_{t_0:t_1} = [s_{t_0}, \ldots, s_{t_1}]$. We use $T_{\text{pred}}$ to denote the prediction horizon and $t_\text{obs}$ to denote the length of the observed history, so that the full trajectory has length $T = t_\text{obs} + T_{\text{pred}}$. Let $\mathcal{D} = \{s^{(i)}_{1:T}\}_{i=1}^{N}$ denote the training dataset of $N$ demonstrated trajectories, where each $s^{(i)}_{1:T} = [s^{(i)}_1, \ldots, s^{(i)}_T]$ is a full trajectory of horizon $T$.

\subsection{Problem Statement}\label{sec:problem}
We aim to develop a data-driven method for generating multimodal predictions of a target agent's future states based on its observed historical states and a dataset of demonstrated trajectories from other agents.
\begin{problem}\label{prob:statement}
    Given the observed trajectory $s_{1:t_\text{obs}}$ of a target agent as input, and a training dataset $\mathcal{D}$ containing full demonstrated trajectories, predict a set of $K$ future trajectory predictions, $\mathcal{T} = \{\hat{s}^k_{t_\text{obs}+1:T}\}_{k=1}^K$, with each candidate trajectory $\hat{s}^k_{t_\text{obs}+1:T}$ representing a plausible behavioral mode.
\end{problem}

The set $\mathcal{T}$ is intended to cover the space of likely behaviors, thus capturing the inherent uncertainty and multimodality of agent motion. We consider two versions of \Cref{prob:statement}:
\begin{enumerate}
\item A minimal, purely data-driven formulation based only on trajectory observations (Section~\ref{sec:base});
\item An extended formulation that uses additional behavioral side information (when available) to constrain the predictions (Section~\ref{sec:extended}).
\end{enumerate}
\vspace{4pt}
\begin{figure}[ht]
    \centering
    \includegraphics[width=1.0\linewidth]{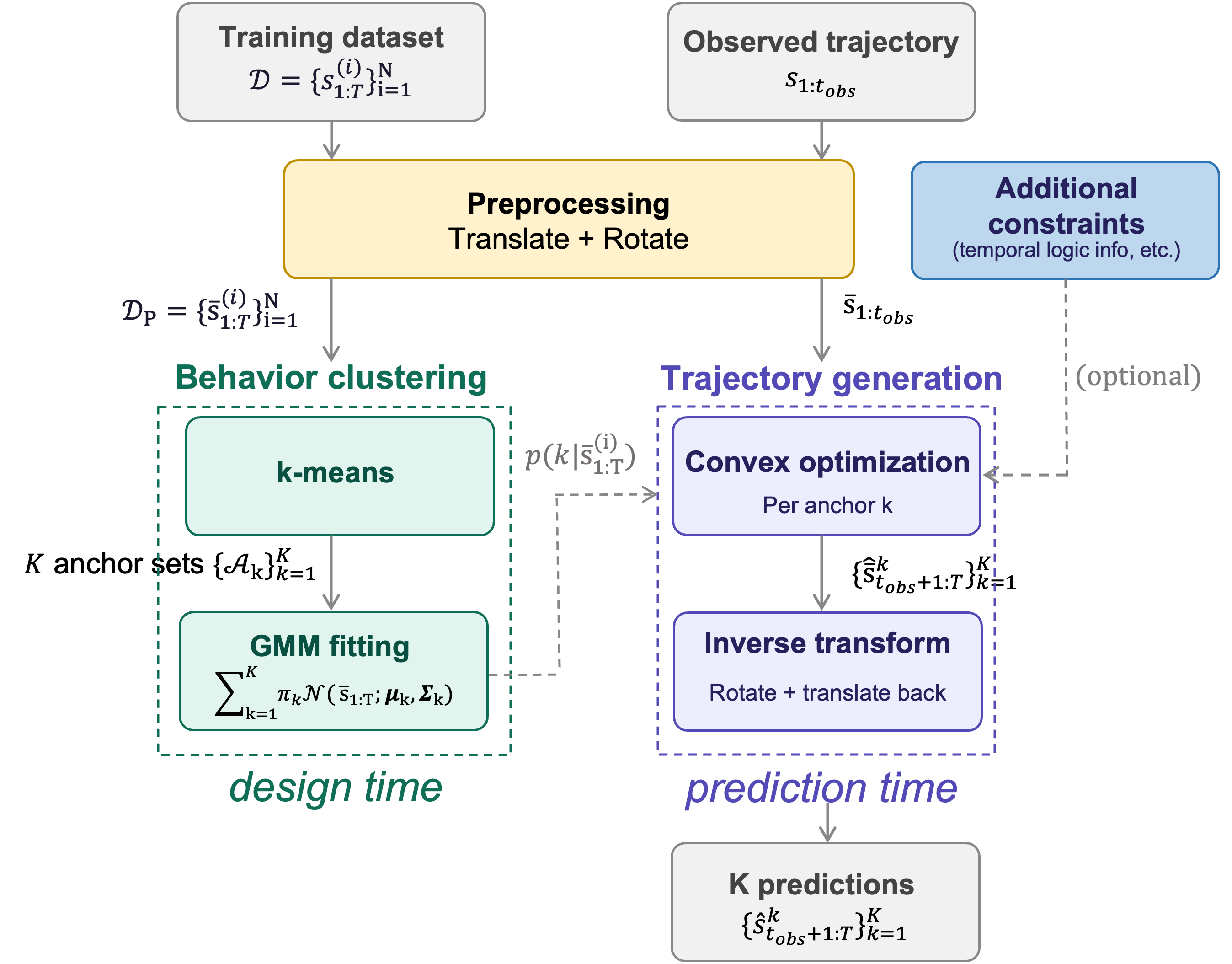}
    \caption{\method{} overview. At design time, training trajectories are preprocessed, clustered, and fitted with a GMM. At prediction time, the observed prefix is preprocessed and one future per anchor is generated by convex optimization, optionally with additional side information.}
    \label{fig:overview}
\end{figure}

\section{\texorpdfstring{\method}{ProMo-CoOp} Base Approach}\label{sec:base}

We propose a multimodal trajectory prediction framework called \method{}. Figure~\ref{fig:overview} shows an overview of the proposed framework. At design time, the training dataset $\mathcal{D}$ is first preprocessed to remove global frame dependencies via translation and rotation (Section~\ref{sec:preprocessing}). The preprocessed trajectories are then partitioned into $K$ anchor sets via $k$-means clustering, and a Gaussian Mixture Model (GMM) is fitted to obtain a probabilistic mode assignment for each training trajectory (Section~\ref{sec:clustering}). At prediction time, the observed partial trajectory is preprocessed using the same procedure. For each anchor $k$, a convex optimization problem is solved to generate a candidate future trajectory as a convex combination of training trajectories, weighted by the precomputed mode probabilities and constrained to remain consistent with the observed partial trajectory (Section~\ref{sec:optim}). The $K$ candidate trajectories are then mapped back to the original coordinate frame via the inverse transform, yielding the final set of predictions $\{\hat{s}^k_{t_\text{obs}+1:T}\}_{k=1}^K$. Since the optimization problem is convex, additional behavioral constraints can be seamlessly incorporated without modifying the clustering stage or requiring retraining; we introduce this as an extended method in Section~\ref{sec:extended}. 
\subsection{Preprocessing}\label{sec:preprocessing}

We preprocess both the training dataset $\mathcal{D}$ and the observed partial trajectory $s_{1:t_\text{obs}}$ of the target agent as follows:
\begin{enumerate*}
\item each trajectory is translated so that the initial observation point $s_1$ is placed at the origin, and
\item the trajectory is rotated to align with a fixed reference orientation. We refer to the resulting coordinate system as the \emph{ego frame}.
\end{enumerate*}

Formally, let $x^{(i)}_{1:T} = s^{(i)}_{1:T} - s^{(i)}_1 \in \mathbb{R}^{T \times 2}$ denote the translated trajectory. Let $Y \in \mathbb{R}^{T \times 2}$ be an arbitrary fixed reference trajectory.\footnote{The choice of $Y$ does not affect the final predictions since the transformation is fully reversed; however, we recommend aligning with a standard axis (e.g., the positive $x$- or $y$-axis) for simplicity.} The optimal rotation matrix $R^{(i)}$ that best aligns $x^{(i)}_{1:T}$ with $Y$ is determined via Procrustes analysis~\cite{gower1975generalized}. Specifically, the cross-covariance matrix $M^{(i)} = Y^\top x^{(i)}_{1:T} \in \mathbb{R}^{2 \times 2}$ is decomposed via singular value decomposition as $M^{(i)} = U^{(i)} \Sigma^{(i)} {V^{(i)}}^\top$, and the rotation matrix is:
\vspace{-0.1cm}
\begin{equation}
R^{(i)} = {V^{(i)}} {U^{(i)}}^\top,
\end{equation}
where the last row of $V^{(i)}$ is negated when $\det(R^{(i)}) < 0$ to ensure that $R^{(i)}$ is a proper rotation (i.e., no reflections). We define the preprocessed trajectory as $\bar{s}^{(i)}_{1:T} = x^{(i)}_{1:T} \, R^{(i)}$, and the preprocessed training dataset is denoted $\mathcal{D}_p = \{\bar{s}^{(i)}_{1:T}\}_{i=1}^{N}$. The observed partial trajectory $s_{1:t_\text{obs}}$ is preprocessed analogously, yielding $\bar{s}_{1:t_\text{obs}}$ with corresponding rotation matrix $R_\text{obs}$.

Since the transformation comprises only a translation and a rotation, it is fully reversible. Given a predicted trajectory $\hat{\bar{s}}^k_{t_\text{obs}+1:T}$ in the ego frame, where $k \in \{1, \ldots, K\}$ indexes the predicted behavioral mode, each predicted state $\hat{\bar{s}}^k_t$ is mapped back to the original coordinate frame via,
\begin{equation}
\hat{s}^k_t = \hat{\bar{s}}^k_t \, R_\text{obs}^{-1} + s_1, \quad t = t_\text{obs}+1, \ldots, T,
\end{equation}
where $s_1$ is the initial observation point of the target agent, yielding the final set of predictions $\{\hat{s}^k_{t_\text{obs}+1:T}\}_{k=1}^K$ in the original coordinate frame. The preprocessed datasets $\mathcal{D}_p$ and $\bar{s}_{1:t_\text{obs}}$ serve as inputs to the behavior clustering and trajectory generation stages described in Sections~\ref{sec:clustering} and~\ref{sec:optim}, respectively.

\subsection{Behavior Clustering}\label{sec:clustering}

To establish a structured representation of multimodal behaviors, we generate $K$ trajectory anchor sets, each capturing a distinct behavioral mode. We apply $k$-means clustering~\cite{1056489} offline to $\mathcal{D}_p$, partitioning it into $K$ anchor sets $\{\mathcal{A}_k\}_{k=1}^K$, where each $\mathcal{A}_k \subseteq \mathcal{D}_p$ contains all trajectories assigned to anchor $k$.

Since $k$-means clustering produces compact, homogeneous clusters, as common in the literature, we model each anchor set $\mathcal{A}_k$ as a unimodal Gaussian distribution,
\vspace{-0.1cm}
\begin{equation}
p(\bar{s}_{1:T} \mid k) = \mathcal{N}(\bar{s}_{1:T};\, \boldsymbol{\mu}_k, \boldsymbol{\Sigma}_k),
\end{equation}
where $(\boldsymbol{\mu}_k, \boldsymbol{\Sigma}_k)$ are the mean vector and covariance matrix estimated from $\mathcal{A}_k$. The full trajectory distribution is then captured by a Gaussian Mixture Model (GMM) over the $K$ anchors,
\vspace{-0.1cm}
\begin{equation}
p(\bar{s}_{1:T}) = \sum_{k=1}^{K} \pi_k \, \mathcal{N}(\bar{s}_{1:T};\, \boldsymbol{\mu}_k, \boldsymbol{\Sigma}_k),
\end{equation}
where $\pi_k = |\mathcal{A}_k| / \sum_{j=1}^{K} |\mathcal{A}_j|$ is the mixing weight of anchor $k$. Given each training trajectory $\bar{s}^{(i)}_{1:T} \in \mathcal{D}_p$, its posterior probability of belonging to behavioral mode $k$ is computed via Bayes' rule,
\vspace{-0.1cm}
\begin{equation}
p(k \mid \bar{s}^{(i)}_{1:T}) = \frac{p(\bar{s}^{(i)}_{1:T} \mid k)\, \pi_k}{\sum_{j=1}^{K} p(\bar{s}^{(i)}_{1:T} \mid j)\, \pi_j},
\end{equation}
where $p(\bar{s}^{(i)}_{1:T} \mid k)$ is the likelihood under anchor $k$ and $\pi_k$ serves as the prior over modes. The resulting mode probabilities $\{p(k \mid \bar{s}^{(i)}_{1:T})\}_{k=1}^K$ serve as soft mode assignments for each training sample, guiding the optimization-based trajectory generation stage described in Section~\ref{sec:optim}.


\subsection{Optimization-Based Trajectory Generation}\label{sec:optim}

Using the posterior mode probabilities $p(k \mid \bar{s}^{(j)}_{1:T})$ of the training samples obtained in the previous stage, we generate $K$ candidate trajectories, one per anchor, by formulating trajectory prediction as a convex optimization problem. The core idea is to represent each predicted trajectory as a convex combination of training samples in $\mathcal{D}_p$, bounding the space of plausible future motions within the convex hull of the training set.

This formulation is supported by an asymptotic theoretical result: the probability that the convex hull formed by $n$ independent samples from an underlying distribution contains a new sample from the same distribution asymptotically converges to $1$ as $n \rightarrow \infty$ \cite{gardes:tel-00540747, Majumdar_2009}. Although this does not provide a finite-sample guarantee, it provides theoretical support for our approach when a sufficiently large set of demonstrations is available.

For each target agent with observed partial trajectory $\bar{s}_{1:t_\text{obs}}$, we generate $K$ candidate trajectories in the ego frame by solving the following convex optimization problem for each~$k$:
\vspace{-0.5cm}
\begin{subequations}\label{eq:gt_pb}
\begin{align}
\underset{w,\epsilon,\hat{\bar{s}}}{\text{minimize}}\; & \sum_{t=1}^{t_\text{obs}} \epsilon_{t} + \sum_{i=1}^{N} \frac{w_i}{p(k \mid \bar{s}^{(i)}_{1:T})+\delta} \label{eq:obj}\\
\text{subject to} \quad & \|\sum_{i=1}^{N} w_i\, C(\bar{s}^{(i)}_t - \bar{s}_t)\|_\infty\leq \epsilon_{t}, \quad t \leq t_{\text{obs}},\label{eq:obs}\\
& \hat{\bar{s}}^k_{t} = \sum_{i=1}^{N} w_i\, \bar{s}^{(i)}_{t}, \quad t_\text{obs}+1\leq t\leq T, \label{eq:convex}\\
& \sum_{i=1}^{N} w_i = 1, \hspace{1.04cm} 0 \leq w_i \leq 1, \label{eq:w}
\end{align}
\end{subequations}
where \(\epsilon_t \geq 0\) is a slack variable that softens the observation-consistency constraint to ensure feasibility, $w_i$ is the weight of $\bar{s}^{(i)}_{1:T}$, $\delta>0$ is a small constant added to the mode posterior probability to avoid division by zero and improve numerical stability, $\bar{s}^{(i)}_t$ is the state at time $t$ of the $i$-th training sample in $\mathcal{D}_p$, $\bar{s}_t$ is the state at time $t\in[1,t_{\text{obs}}]$ of the target agent, $\hat{\bar{s}}_{t}^k$ is the predicted state at time $t\in[t_{\text{obs}}+1,T]$ under anchor $k$ of the target agent. The matrix $C \in \mathbb{R}^{4 \times 2}$ defines a zonotopic norm approximating the Euclidean ball using four directions,
\begin{equation}
C = \begin{bmatrix} 1 & 0 & \frac{\sqrt{2}-1}{\sqrt{2}} & \frac{\sqrt{2}-1}{\sqrt{2}}\\[4pt] 0 & 1 & \frac{\sqrt{2}-1}{\sqrt{2}} & -\frac{\sqrt{2}-1}{\sqrt{2}} \end{bmatrix}^\top.
\end{equation}

The optimization enforces three requirements:
\begin{enumerate*}[label=\it{(\roman*)}]
\item \emph{convex hull adherence}~\eqref{eq:convex},\eqref{eq:w}: $\hat{\bar{s}}^k_t$ is constrained to lie in the convex hull of the training trajectories in $\mathcal{D}_p$;
\item \emph{observation consistency}~\eqref{eq:obs}: the zonotopic norm $\|\sum_{i=1}^{N} w_i\, C(\bar{s}^{(i)}_t - \bar{s}_t)\|_\infty\leq \epsilon_{t}$ bounds the deviation from the observed states $\bar{s}_{1:t_\text{obs}}$, providing a tractable polyhedral relaxation of the Euclidean distance constraint;\footnote{In our experiments, the zonotopic norm defined by $C$ yielded slightly better results than the standard $\ell_1$ norm, though the difference was not significant; other polyhedral norms may be used interchangeably.} and
\item \emph{behavioral mode alignment}~\eqref{eq:obj}: the term $w_i / (p(k \mid \bar{s}^{(i)}_{1:T})+\delta)$ penalizes weights assigned to training samples inconsistent with anchor $k$, encouraging the predicted trajectory to conform to the target behavioral mode.
\end{enumerate*}
After solving the optimization for each anchor $k$, the predicted trajectories $\{\hat{\bar{s}}^k_{t_\text{obs}+1:T}\}_{k=1}^K$ in the ego frame are mapped back to the original coordinate frame via the inverse transform described in Section~\ref{sec:preprocessing}, yielding the final candidate set $\mathcal{T} = \{\hat{s}^k_{t_\text{obs}+1:T}\}_{k=1}^K$.


\section{STL-Based Extension}\label{sec:extended}

A key advantage of the optimization-based formulation in Section~\ref{sec:optim} is its flexibility to incorporate additional convex constraints beyond the three core requirements \eqref{eq:obs}--\eqref{eq:w}, without modifying the behavior clustering stage or requiring retraining, preserving the modularity of the framework. Formally, the extended optimization problem augments~\eqref{eq:gt_pb} with an additional set of convex constraints $\mathcal{C}$,
\begin{subequations}\label{eq:extended_pb}
\begin{align}
\underset{w,\epsilon,\hat{\bar{s}}}{\text{minimize}} \quad & \sum_{t=1}^{t_\text{obs}} \epsilon_{t} + \sum_{i=1}^{N} \frac{w_i}{p(k \mid \bar{s}^{(i)}_{1:T})+\delta} \\
\text{subject to} \quad & \eqref{eq:obs}, \eqref{eq:convex}, \eqref{eq:w}, \nonumber\\
& f_c(\hat{\bar{s}}^k_t) \leq 0, \quad c \in \mathcal{C}, \quad t_\text{obs}+1\leq t\leq T,
\end{align}
\end{subequations}
where $f_c$ denotes a convex constraint function. This general formulation can accommodate a wide range of constraints by expressing them as convex constraints or convex approximations, including physical constraints such as kinematic limits and speed bounds, as well as scene-level constraints such as collision avoidance and road boundary compliance. In this work, we apply this framework using behavioral constraints derived from Signal Temporal Logic (STL)~\cite{maler2004monitoring, fainekos2009robustness}, a formal language for specifying properties over real-valued signals that provides interpretable and mathematically rigorous behavioral rules inferred directly from data. The predicate satisfaction probabilities required by these constraints are estimated using Conditional Flow Matching (CFM), described in Section~\ref{sec:stl_constraints}.

\subsection{Signal Temporal Logic}\label{sec:stl}

We leverage STL~\cite{maler2004monitoring, fainekos2009robustness} to formally characterize the spatiotemporal properties of each behavioral mode. We consider a fragment of STL with the syntax,
\begin{equation}\label{eq:stl-frag}
\phi ::= \top \mid \mu \mid \neg\phi \mid \phi_1 \land \phi_2 \mid \phi_1 \lor \phi_2 \mid F_{[t_1,t_2]}\phi \mid G_{[t_1,t_2]}\phi,
\end{equation}
where $\mu := a^\top \bar{s}_t \geq b$, $a \in \mathbb{R}^2$, $b \in \mathbb{R}$, is a linear predicate over the agent state in the ego frame. The Boolean operators $\neg$, $\land$, $\lor$ denote negation, conjunction, and disjunction, respectively. The temporal operators $F_{[t_1,t_2]}\phi$ and $G_{[t_1,t_2]}\phi$ denote \emph{eventually} and \emph{always}: $F_{[t_1,t_2]}\phi$ holds if $\phi$ is satisfied at least once in $[t_1,t_2] \cap \mathbb{Z}$, and $G_{[t_1,t_2]}\phi$ holds if $\phi$ is satisfied at all time steps in $[t_1,t_2] \cap \mathbb{Z}$.
\subsection{STL-Guided Behavioral Constraints}\label{sec:stl_constraints}
\noindent \textbf{Inferring STL formulas.}\quad To formally characterize the behavioral patterns of each anchor, we infer a unique STL formula $\phi_k$ directly from the anchor set $\mathcal{A}_k$ using TLINet~\cite{li2023learning, li2024tlinet}, a differentiable neural network framework for temporal logic inference. The inferred formula set $\Phi = \{\phi_k \mid k = 1,\ldots,K\}$ satisfies $\bar{s}_{1:T} \models \phi_k$ if and only if $\bar{s}_{1:T} \in \mathcal{A}_k$, ensuring that each formula distinctly captures the semantic properties of its associated anchor in a human-readable form, such as ``turn left'' or ``maintain speed''.

\noindent \textbf{Predicate satisfaction estimation.}\quad We define the predicate set $\mathcal{M} = \{\mu_{k,j} \mid k = 1,\ldots,K;\, j = 1,\ldots,N_k\}$, where $\mu_{k,j} := a_{k,j}^\top \bar{s}_t \geq b_{k,j}$ is the $j$-th predicate of $\phi_k$ and $N_k$ is the number of predicates in $\phi_k$. To estimate the probability of satisfaction of each predicate at future time steps, we train a CFM model~\cite{lipman2023flowmatching} that learns the conditional distribution of the full future trajectory segment $\bar{s}_{t_\text{obs}+1:T}$ given the observed prefix $\bar{s}_{1:t_\text{obs}}$. The model comprises a convolutional signal encoder $f_\phi$ that maps the observed prefix to a context embedding $z = f_\phi(\bar{s}_{1:t_\text{obs}}) \in \mathbb{R}^{d_z}$, and a vector field network $v_\theta(\bar{s}_t, t, z)$ trained by regressing toward a conditional velocity field,
\begin{equation}
\mathcal{L}_\text{CFM} = \mathbb{E}_{t, \bar{s}_0, \bar{s}_1} \left[ \|v_\theta(\bar{s}_t, t, z) - v_t(\bar{s}_t \mid \bar{s}_1)\|^2 \right],
\end{equation}
where $\bar{s}_1 \in \mathbb{R}^{2T_{\text{pred}}}$ is the flattened future trajectory, $\bar{s}_0 \sim p_0 = \mathcal{N}(0, \sigma^2 I)$, $\bar{s}_t = (1-t)\bar{s}_0 + t\bar{s}_1$, and $v_t(\bar{s}_t \mid \bar{s}_1) = \bar{s}_1 - \bar{s}_0$. Both components are trained jointly using the Adam optimizer. When predicting for a target agent, $M$ future trajectory samples $\{\bar{s}^{(m)}_{t_\text{obs}+1:T}\}_{m=1}^M$ are obtained by integrating $v_\theta$ from $t=0$ to $t=1$ using an Euler solver, and the predicate satisfaction probability at each time step $t$ is estimated as,
\begin{equation}
p^{k,j}_{t} = \frac{1}{M} \sum_{m=1}^{M} \mathbf{1}\left[ a_{k,j}^\top \bar{s}^{(m)}_{t} - b_{k,j} \geq 0 \right].
\end{equation}
\noindent \textbf{STL-constrained trajectory generation.}\quad The CFM-estimated probabilities $\{p^{k,j}_{t}\}$ are incorporated into the following extended convex optimization problem:
\begin{subequations}\label{eq:stl_pb}
\begin{align}
\underset{w,\epsilon,\gamma,\hat{\bar{s}}}{\text{minimize}}  \quad & \sum_{t=1}^{t_\text{obs}} \epsilon_{t} + \sum_{i=1}^{N} \frac{w_i}{p(k \mid \bar{s}^{(i)}_{1:T})+\delta} + \sum_{t=t_\text{obs}+1}^{T}\sum_{j=1}^{N_k} \gamma_{j,t} \label{eq:stl_obj}\\
\text{subject to} \quad & \eqref{eq:obs}, \eqref{eq:convex}, \eqref{eq:w}, \nonumber\\
& \mu_{k,j}(\hat{\bar{s}}^k_{t}) \geq -\gamma_{j,t}(1 - p^{k,j}_{t}+\delta), \quad \text{if } p^{k,j}_{t} \geq 0.5, \nonumber\\
& \quad t_\text{obs}+1\leq t\leq T \quad \forall j = 1,\ldots,N_k, \label{eq:stl_pos}\\
& \mu_{k,j}(\hat{\bar{s}}^k_{t}) \leq \gamma_{j,t}\, (p^{k,j}_{t}+\delta), \quad \text{if } p^{k,j}_{t} < 0.5, \nonumber\\
& \quad t_\text{obs}+1\leq t\leq T \quad \forall j = 1,\ldots,N_k, \label{eq:stl_neg}
\end{align}
\end{subequations}
where \(\gamma_{j,t} \geq 0\) are slack variables that soften the predicate constraints, allowing constraint violations when necessary and preserving feasibility of the extended optimization, and $\mu_{k,j}(\bar{s}^k_{t}) = a_{k,j}^\top \hat{\bar{s}}^k_{t} - b_{k,j}$ denotes the signed robustness value of predicate $\mu_{k,j}$ at the predicted state. The soft formulation in~\eqref{eq:stl_pos}--\eqref{eq:stl_neg} biases trajectory generation according to the predicted satisfaction pattern, while preserving the convexity of the optimization problem.

The STL-derived constraints provide a mechanism for incorporating interpretable behavioral side information into trajectory generation beyond purely statistical mode assignment. The contribution of STL-derived constraints relative to the base method is evaluated in Section~\ref{sec:exp_stl}.

\section{Experiments}
This section evaluates the ProMo-CoOp base approach and its STL-based extension on the CARLA Town01 dataset. The experiments focus on the single-agent setting and are intended to evaluate the effects of probabilistic mode identification, constrained trajectory generation, and STL-derived behavioral side information within this setting. Section~\ref{sec:exp_base} compares the \method{} base approach against representative trajectory-prediction baselines, including both a simple deterministic baseline and strong multimodal predictors. Section~\ref{sec:exp_stl} then evaluates the STL extension under both known and inferred constraint information, together with related ablations.
\subsection{Dataset}
We evaluate our framework on the CARLA dataset introduced by Rhinehart et al.~\cite{rhinehart2019precog}, generated using the CARLA autonomous driving simulator~\cite{dosovitskiy2017carla}. The dataset was collected by running an autopilot in Town01 for over 900 episodes of 100 seconds each, recording the trajectory of every vehicle at 5 Hz. It comprises 60,701 training, 7,586 validation, and 7,567 test scenes. We use the single-agent setting ($A=1$), where the task is to predict the future trajectory of one target vehicle over a 4-second horizon (20 time steps) given an initial 2-second observation (10 time steps).
\subsection{Baselines}
We compare our framework against the following trajectory prediction methods.

\textbf{DESIRE}~\cite{lee2017desire} generates a diverse set of trajectory hypotheses by sampling from a Conditional Variational Autoencoder (CVAE), and subsequently ranks and refines them through a deep Inverse Optimal Control (IOC) network that incorporates both static scene context and dynamic agent interactions.

\textbf{Social GAN}~\cite{gupta2018social} combines a recurrent sequence-to-sequence generator with a generative adversarial framework to produce socially plausible multimodal predictions. A novel pooling mechanism aggregates information across neighboring agents, and prediction diversity is encouraged through a variety loss during adversarial training.

\textbf{PRECOG}~\cite{rhinehart2019precog} presents a probabilistic generative model for trajectory prediction, learning a flow-based distribution over future trajectories conditioned on past positions and scene observations.

\textbf{Conditional Flow Matching (CFM)}~\cite{lipman2023flowmatching} is used here as a purely generative baseline that models the conditional trajectory distribution without explicit mode decomposition or constraint enforcement.

\textbf{Linear}~\cite{becker2018red} is a simple deterministic baseline that extrapolates future positions by fitting a linear model to the observed trajectory, assuming constant velocity throughout the prediction horizon.
\subsection{Implementation Details}
We utilize only the 2D position coordinates of the target agent. For the Procrustes analysis, we use a unit vector pointing east (along the positive $x$-axis) as the fixed reference orientation to align the agent's general heading.

For $k$-means clustering, we adopt the $k$-means++ initialization strategy~\cite{arthur2006k} to improve centroid initialization and reduce sensitivity to random seed selection. The optimal number of clusters $K$ is determined by maximizing the silhouette score~\cite{rousseeuw1987silhouettes} on the training set, selecting from the candidate set $K \in \{6, 10, 12, 15\}$. This yields $K=12$ for the CARLA dataset.

The optimization is solved using Gurobi. For the CFM-based predicate satisfaction predictor, the satisfaction probability $p^{k,i}_{t}$ is estimated using $M=500$ Monte Carlo samples drawn from the learned flow. Our CFM implementation draws on the TorchCFM library and related code release of Tong et al.~\cite{tong2024improving, tong2023simulation}.
\subsection{Evaluation Metrics}

We evaluate the methods using the following three metrics, which are standard in multimodal trajectory prediction.

\noindent \textbf{Minimum Mean Squared Displacement (minMSD).}\quad Given $K$ predicted trajectories, minMSD measures the mean squared displacement between the ground-truth future trajectory and the closest predicted trajectory,
\begin{equation}
\text{minMSD} = \min_{k=1,\ldots,K} \frac{1}{T_{\text{pred}}} \sum_{t=t_{obs}+1}^{T} \| \hat{s}^k_t - s^*_t \|^2,
\end{equation}
where $s^*_t$ denotes the ground-truth state at time $t$ and $T_{\text{pred}} = T - t_{obs}$ is the prediction horizon length.

\noindent \textbf{Minimum Final Displacement Error (minFDE).}\quad minFDE measures the Euclidean distance between the ground-truth final position and the endpoint of the closest predicted trajectory,
\begin{equation}
\text{minFDE} = \min_{k=1,\ldots,K} \| \hat{s}^k_T - s^*_T \|,
\end{equation}
where $\hat{s}^k_T$ and $s^*_T$ denote the predicted and ground-truth positions at the final time step $T$, respectively.

\noindent \textbf{Miss Rate (MR).}\quad Miss rate~\cite{chang2019argoverse} measures the proportion of test cases in which all $K$ predicted trajectories fail to come within a distance of $d$ meters of the ground-truth final position,
\begin{equation}
\text{MR}_d = \frac{1}{|\mathcal{D}_{\text{test}}|} \sum_{n=1}^{|\mathcal{D}_{\text{test}}|} \mathbf{1}\left[ \min_{k=1,\ldots,K} \| \hat{s}^{k,n}_T - s^{*,n}_T \| > d \right],
\end{equation}
where $\mathbf{1}[\cdot]$ is the indicator function. A lower miss rate indicates that the predicted set covers the ground-truth final position more reliably.

\subsection{Evaluation of the \texorpdfstring{\method}{ProMo-CoOp} Base Approach}\label{sec:exp_base}
\Cref{tab:carla_results} reports the minMSD at $K=12$ for all methods on the CARLA dataset. Since the source code for several baselines is not publicly available, we were only able to obtain minMSD results for all methods, as this is the metric reported consistently across all baselines. Our framework achieves the lowest minMSD of $0.36$, outperforming all baselines.

The Linear baseline performs the weakest with a minMSD of $2.47$, confirming that the constant velocity assumption is inadequate for capturing the complexity of vehicle motion. Among learning-based methods, DESIRE, SocialGAN and CFM  achieve minMSD scores of $1.07$, $0.92$ and $0.79$ respectively, while PRECOG achieves the second-best performance at $0.50$. Our method surpasses PRECOG by $28\%$, using only 2D position coordinates, demonstrating that the proposed behavior clustering and convex optimization formulation provides a more effective approach to multimodal trajectory prediction without relying on additional sensory input.



\begin{figure}[ht]
    \centering
    \includegraphics[trim=0 8cm 0 3cm, clip, width=\linewidth]{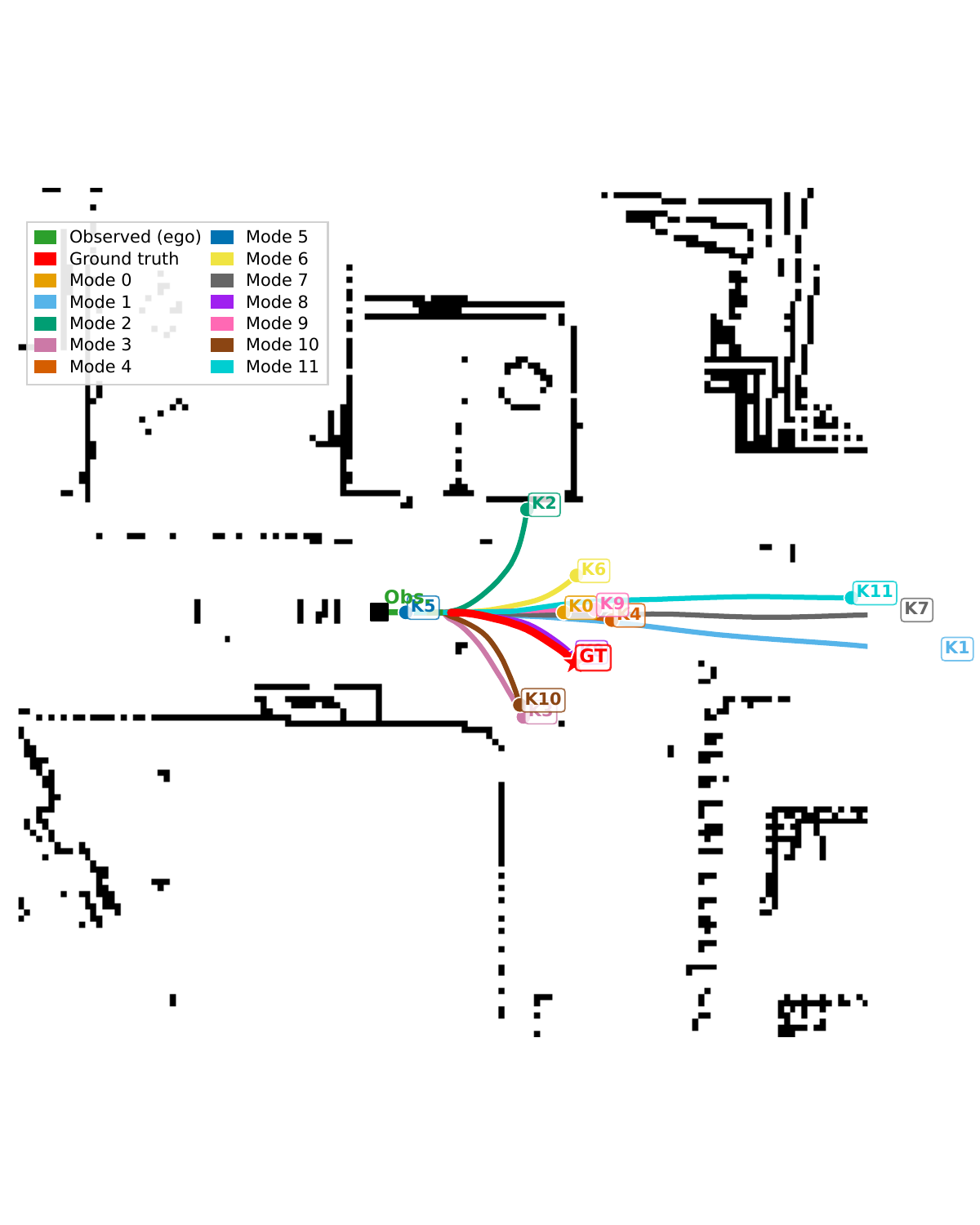}
    \caption{Example of the $K=12$ mode-conditioned trajectory predictions generated by \method{} for a single observed trajectory prefix at an intersection. Each colored curve corresponds to one candidate future associated with a different anchor.}
    \label{fig:carla_result}
\end{figure}

Figure~\ref{fig:carla_result} shows a qualitative example of the $K=12$ candidate trajectories in $\mathcal{T}$ for one observed prefix. The predicted set covers multiple distinct futures associated with different behavioral modes, indicating that the GMM-based clustering in trajectory space captures a significant portion of behavioral variability.
\begin{table}[ht]
\centering
\begin{tabular}{l c}
\toprule
Method & minMSD \\
\midrule
Linear & $2.47$ \\
DESIRE~\cite{lee2017desire} (K=12)  & $1.07$ \\
SocialGAN~\cite{gupta2018social} (K=12) & $0.92$ \\
CFM (K=12) & $0.79$ \\
PRECOG~\cite{rhinehart2019precog} (K=12) & $0.50$ \\
\textbf{Ours} (K=12) & $\mathbf{0.36}$ \\
\bottomrule
\end{tabular}
\caption{Evaluation of \method{} (base) and baselines on the CARLA Dataset.}
\label{tab:carla_results}
\end{table}
\begin{table}[t]
\centering
\caption{Ablation results for the STL extension and related variants on the CARLA dataset.}
\label{tab:ablation_results}
\begin{tabular}{lccc}
\toprule
Method & minMSD & minFDE & $\text{MR}_2$ \\
\midrule
\method{} (base) & 0.36 & 0.65 & 0.06 \\
base + GT STL  & 0.20 & 0.45 & 0.03 \\
base + inferred STL  & 0.48 & 0.68 & 0.08 \\
base w/o GMM & 1.29 & 1.46 & 0.22 \\
base + GT & 0.00 & 0.00 & 0.00 \\
\bottomrule
\end{tabular}
\end{table}
\subsection{Evaluation of the STL Extension and Related Ablations}\label{sec:exp_stl}
We use this set of experiments to evaluate the effect of behavioral side information in the proposed framework. We first evaluate the STL extension using ground-truth predicate satisfactions to isolate the effect of accurate STL-based constraints. We then consider the fully data-driven setting, where predicate satisfactions are inferred from the observed trajectory. Finally, we perform ablations to evaluate the roles of probabilistic mode identification and the representational capacity of the optimization framework. The results are shown in Table~\ref{tab:ablation_results}.

(i) \emph{Ground-truth STL satisfactions (GT STL).} This variant isolates the effect of accurate STL-based behavioral side information. The STL formulas are inferred from data as described in Section~\ref{sec:stl_constraints}, while the ground-truth predicate satisfactions are used in place of the CFM-estimated satisfactions in~\eqref{eq:stl_pos}--\eqref{eq:stl_neg}. Compared with the base method, minMSD decreases from $0.36$ to $0.20$, minFDE decreases from $0.65$ to $0.45$, and MR$_2$ decreases from $0.06$ to $0.03$. These results show that accurate STL-based behavioral constraints can improve trajectory generation within the proposed optimization framework.

(ii) \emph{Inferred STL constraints.} We next consider the fully data-driven setting, where predicate satisfaction probabilities are estimated from the observed trajectory using the CFM-based predictor and incorporated through the soft constraints in~\eqref{eq:stl_pos}--\eqref{eq:stl_neg}. In this setting, minMSD increases from $0.36$ to $0.48$, minFDE increases from $0.65$ to $0.68$, and MR$_2$ increases from $0.06$ to $0.08$. Thus, the inferred STL constraints do not improve the base method in the current experiment. The difference between the GT-STL and inferred-STL results indicates that the main limitation is the accuracy of the predicted predicate satisfactions. Incorrect satisfaction estimates can introduce constraints that are inconsistent with the target agent's actual future behavior, offsetting the benefit obtained from accurate behavioral side information.

(iii) \emph{No GMM.} This ablation isolates the contribution of probabilistic mode identification in the base framework. Starting from the base optimization problem~\eqref{eq:gt_pb}, we remove the behavioral mode alignment term in the objective, i.e., the term $\sum_{j=1}^{N} w_j/(p(k \mid \bar{s}^{(j)}_{1:T})+\delta)$ in~\eqref{eq:obj}. The optimization therefore retains only observation consistency and convex-hull generation, but no longer biases the solution toward a selected behavioral mode. The resulting degradation in performance shows that probabilistic mode identification is important for maintaining distinct multimodal predictions and improving coverage over plausible futures.

(iv) \emph{Ground-truth (GT).} This variant serves as a sanity check on the representational capacity of the optimization framework. We assume the ground-truth future trajectory in the ego frame, denoted by $\bar{s}^{\ast}_{t_\text{obs}+1:T}$, is available, and add the following soft future-matching constraints to the base formulation:
\begin{equation}
\Bigl\|\sum_{i=1}^{N} w_i\, C\!\left(\bar{s}^{(i)}_t-\bar{s}^{\ast}_t\right)\Bigr\|_\infty \leq \epsilon_t,
\quad t_\text{obs}+1 \leq t \leq T.
\end{equation}
This experiment tests whether the convex-combination representation induced by the training set is expressive enough to recover the true future when perfect future information is available. The results show zero error in the reported metrics, indicating that the training set provides sufficient coverage of the evaluated trajectory space to represent the ground-truth futures.
\section{Conclusion}
We presented \method{}, a data-driven framework for multimodal trajectory prediction that separates behavioral mode identification from trajectory generation. Behavioral modes are extracted from data through clustering and probabilistic modeling, while future trajectories are generated through a convex optimization problem that enforces consistency with both the observed history and the selected mode.

This decomposition addresses the three challenges highlighted in the introduction. Probabilistic mode identification improves \emph{coverage} by representing multiple plausible futures and improves \emph{explainability} by making the underlying behavioral modes explicit. The optimization layer improves \emph{feasibility} by constraining predictions to the convex hull of demonstrated trajectories, encouraging them to follow demonstrated dynamics. When reliable STL behavioral side information is available, it can be incorporated within the same framework to further improve feasibility and interpretability.

Empirically, the base method is competitive with strong multimodal baselines, indicating that explicit mode decomposition combined with constrained generation is an effective approach to multimodal trajectory prediction. The ablation results further show that probabilistic mode identification is central to performance, while accurate STL-based behavioral constraints can provide additional improvements. However, the performance of the STL extension depends on the accuracy of the predicate satisfaction estimates, as reflected by the gap between the inferred-STL and GT-STL results. The framework also relies on behavioral modes identified from the training data, so previously unseen behaviors that differ substantially from the demonstrated trajectories may not be well represented by the learned anchors.

Future work will focus on improving predicate-satisfaction prediction, adapting the mode representation to previously unseen behaviors, and extending the framework to real-world datasets, multi-agent settings, and richer scene contexts.




\bibliographystyle{IEEEtran}
\bibliography{2026-cdc-TP}

\end{document}